\documentclass[letterpaper, 10 pt, conference]{ieeeconf}  

\IEEEoverridecommandlockouts                              

\usepackage[caption=false]{subfig}  
\usepackage{hyperref}
\usepackage{array}
\usepackage{graphics} 
\usepackage{epsfig} 
\usepackage{mathptmx} 
\usepackage{times} 
\usepackage{amsmath} 
\usepackage{amssymb}  
\usepackage[caption=false]{subfig}  
\usepackage{color}
\usepackage{xcolor}
\usepackage{multirow}
\usepackage{makecell}
\usepackage{cite}
\usepackage{hyperref}
\hypersetup{colorlinks}
\let\labelindent\relax
\usepackage{enumitem}
\usepackage{dsfont}
\usepackage{xurl}
\usepackage{graphicx}  
\usepackage{cuted}     
\usepackage{caption}   

\IEEEaftertitletext{\vspace{-24mm}}  

\title{\LARGE \bf
RGB-SQ Grasp: Inferring Local Superquadric Primitives from Single RGB Image for Graspability-Aware Bin Picking
}

\author{Yifeng Xu$^{1,2}$, Fan Zhu$^3$, Ye Li$^1$, Sebastian Ren$^4$, Xiaonan Huang$^1$, Yuhao Chen$^2$ 
\thanks{$^1$Yifeng Xu, Ye Li, and Xiaonan Huang are with University of Michigan, Ann Arbor, MI, 48109, USA. {\tt\small \{yifengxu, yeyli, xiaonanh\}@umich.edu}}
\thanks{$^2$Yuhao Chen is with the University of Waterloo, Waterloo, ON, N2L 3G1, Canada. {\tt\small yuhao.chen1@uwaterloo.ca}}
\thanks{$^3$Fan Zhu is with the Xi'an Jiaotong-Liverpool University, Suzhou, Jiangsu, 215123, China. {\tt\small Fan.Zhu@xjtlu.edu.cn}}
\thanks{$^4$Sebastian Ren is with the University College London, London, WC1E 6BT, UK. {\tt\small zcabsre@ucl.ac.uk}}
}

\begin{document}
\maketitle
\begin{strip}
  \centering
  \includegraphics[width=1.0\textwidth]{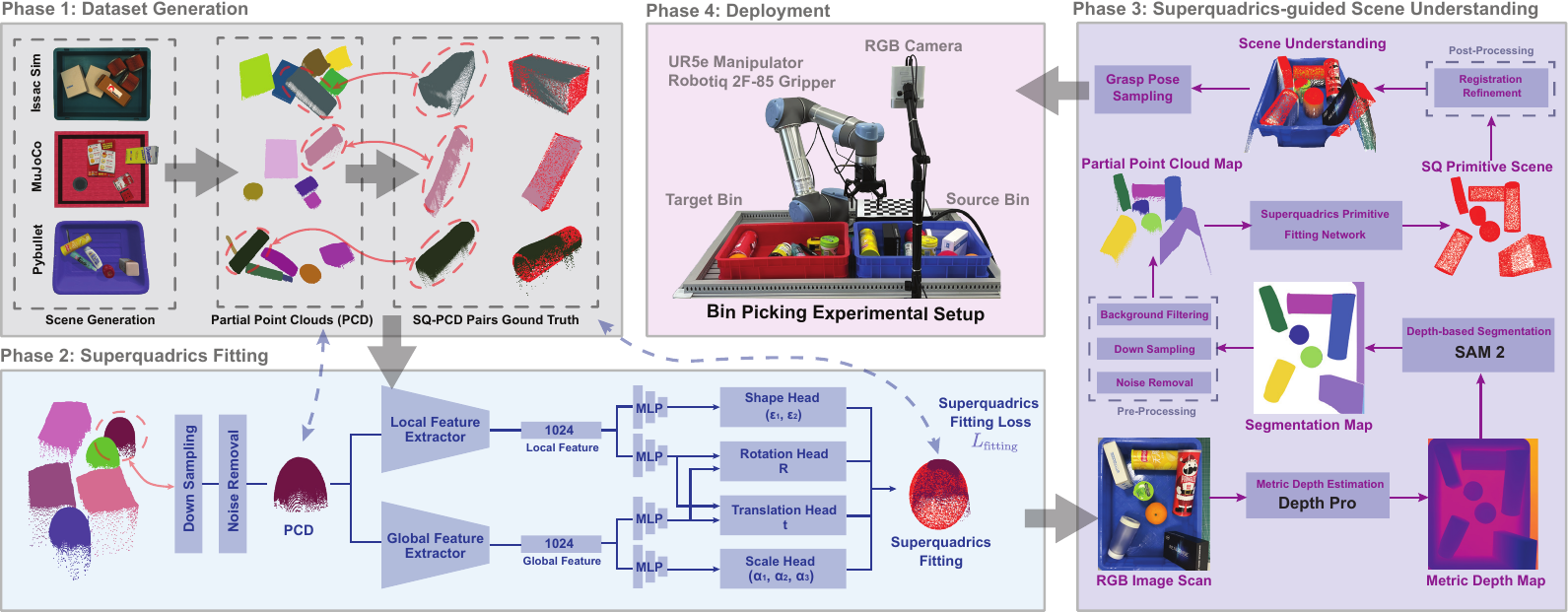}
  \captionof{figure}{\textbf{Overview of the RGBSQGrasp framework.} (1) Dataset generation using cross-platform simulators to create partial point clouds and superquadric ground truth pairs, (2) Superquadric fitting network with local and global feature extraction, (3) Object scene understanding using RGB image scan, depth estimation, and superquadric primitive fitting, and (4) Deployment in a real bin-picking experimental setup with a UR5e manipulator and RGB camera.}
  \label{fig:RGBSQGrasp}
  \vspace{0.5em}
\end{strip}
\thispagestyle{empty}
\pagestyle{empty}

\begin{abstract}
Superquadrics (SQ) offer a compact, interpretable shape representation that captures the physical and graspability understanding of objects. In this work, we propose RGBSQGrasp, a grasping framework that leverages superquadric shape primitives and foundation metric depth estimation models to infer grasp poses from a monocular RGB camera—eliminating the need for depth sensors. Our framework integrates a cross-platform dataset generation pipeline, a foundation model-based object point cloud estimation module, a global-local superquadric fitting network, and an SQ-guided grasp pose sampling module. By integrating these components, RGBSQGrasp reliably infers grasp poses through geometric reasoning. Real-world robotic experiments demonstrate a 92\% grasp success rate, highlighting the effectiveness of RGBSQGrasp in packed bin-picking environments. Supplementary demonstration videos can be found at project website: \href{https://rgbsqgrasp.github.io/}{https://rgbsqgrasp.github.io/}


\end{abstract}


\section{INTRODUCTION}

In industrial environments such as warehousing and manufacturing, robotic bin picking is crucial for tasks like inventory management and assembly line operations \cite{BinPickingBenchmark}. However, objects in unstructured bin-picking environments are often randomly placed, overlapping, or entangled, complicating these tasks. Unlike tabletop settings, where objects are typically isolated and positioned in a more predictable manner, bin environments present two major challenges: (1) the available camera views are limited, making it difficult to reconstruct a complete object model through multi-view fusion, and (2) tight spatial constraints and occlusions caused by object-to-object and object-to-bin interactions, which lead to failed recognition and grasping of objects \cite{symmetry1}. These challenges limit grasping effectiveness and make it difficult for robots to manipulate objects in cluttered environments \cite{Li2022}. To address these challenges, techniques capable of inferring grasping actions from partial or incomplete data, such as partial point clouds from limited perspectives, are essential for bin-picking systems \cite{BinPickingBenchmark,GraspPoseDetection}.


Classical approaches to robotic grasping can be broadly categorized into two types: (1) Analytical approaches, which rely on the physical models of objects to perform force-closure or form-closure analysis \cite{{Roa2014}, {Bicchi2000}, {TRO1}}. Stable grasp contact points are planned under the assumption that the object’s geometry and physics are fully known. This approach offers intrinsic descriptiveness of objects and scenes, utilizing a proper understanding of physics to ensure grasp stability \cite{Bicchi2000, TRO1}. However, the need of often-inaccessible object models limits its generalization to novel objects.
\cite{graspit2023}. Data-driven approaches have emerged to overcome the limitations of analytical methods \cite{{grasp_synthesis_review}}. These approaches generate grasp hypotheses by training models on diverse datasets with labeled grasp poses, thereby improving generalization. However, high-quality datasets used to train these models are often difficult to acquire. To mitigate this issue, numerous studies have utilized synthetic data to simulate real-world scenarios \cite{symmetry3,lin2020,Fei2022}. Nonetheless, a significant sim-to-real gap remains, resulting in inconsistencies when the models are deployed on real robots\cite{{lin2020}}. 

The use of shape primitives originated in analytical approaches, based on the observation that everyday objects can be represented by geometric primitives \cite{{Miller2003}, {vezzani2017}}. Among these, superquadric primitives have gained increasing interest in robotic grasping due to their compact parametrization and generalization to novel shapes, offering advantages in both object understanding and grasp evaluation \cite{ {vezzani2017}, {symmetry2}, {goldfeder2007}, {super5}, {vezzani2018}}. Superquadric fitting is typically framed as an optimization problem \cite{{Liu2022}}, requiring nearly complete point clouds input, which is impractical in settings with limited camera views, such as bin picking. Previous work has explored using depth images for primitive recognition due to their richer geometric information \cite{lin2020, access2024}. However, these methods often depend on high-quality depth sensors, as noise and discrepancies from real-world sensors degrade performance \cite{access2024}, thereby introducing expensive resource dependencies \cite{grasp_synthesis_review}.

To address these challenges, we propose a superquadric-guided grasping framework for bin picking using a single RGB sensor. The framework comprises a universal, cross-platform synthetic dataset generation pipeline using physics engines to simulate real-world scene complexities, a data-driven superquadric fitting network to infer superquadrics from locally visible geometry, and an inference module that leverages state-of-the-art vision foundation models for enhanced generalization and robustness  \cite{foundation0, foundation2}. Our key contributions are as follows:

\begin{itemize}[leftmargin=*]
\item We establish a superquadric-guided framework for bin picking in limited-view environments, using a single monocular RGB image to infer superquadric representations from locally visible geometry without requiring prior object knowledge or multi-camera views.
\item We introduce a data-driven superquadric fitting method that integrates local and global feature extraction. This approach achieves a 92\% success rate in packed bin-picking scenes, improving grasp success by 11.5\% over the baseline by leveraging superquadric primitives for efficient grasp synthesis and evaluation.
\item Leveraging vision foundation models, our approach is independent of depth sensor type and quality, ensuring robustness and consistency in real-world deployments.
\end{itemize}

\section{RELATED WORK}

\subsection{Preliminary}

\begin{figure}[!t]
    \centering
    \includegraphics[scale=0.60]{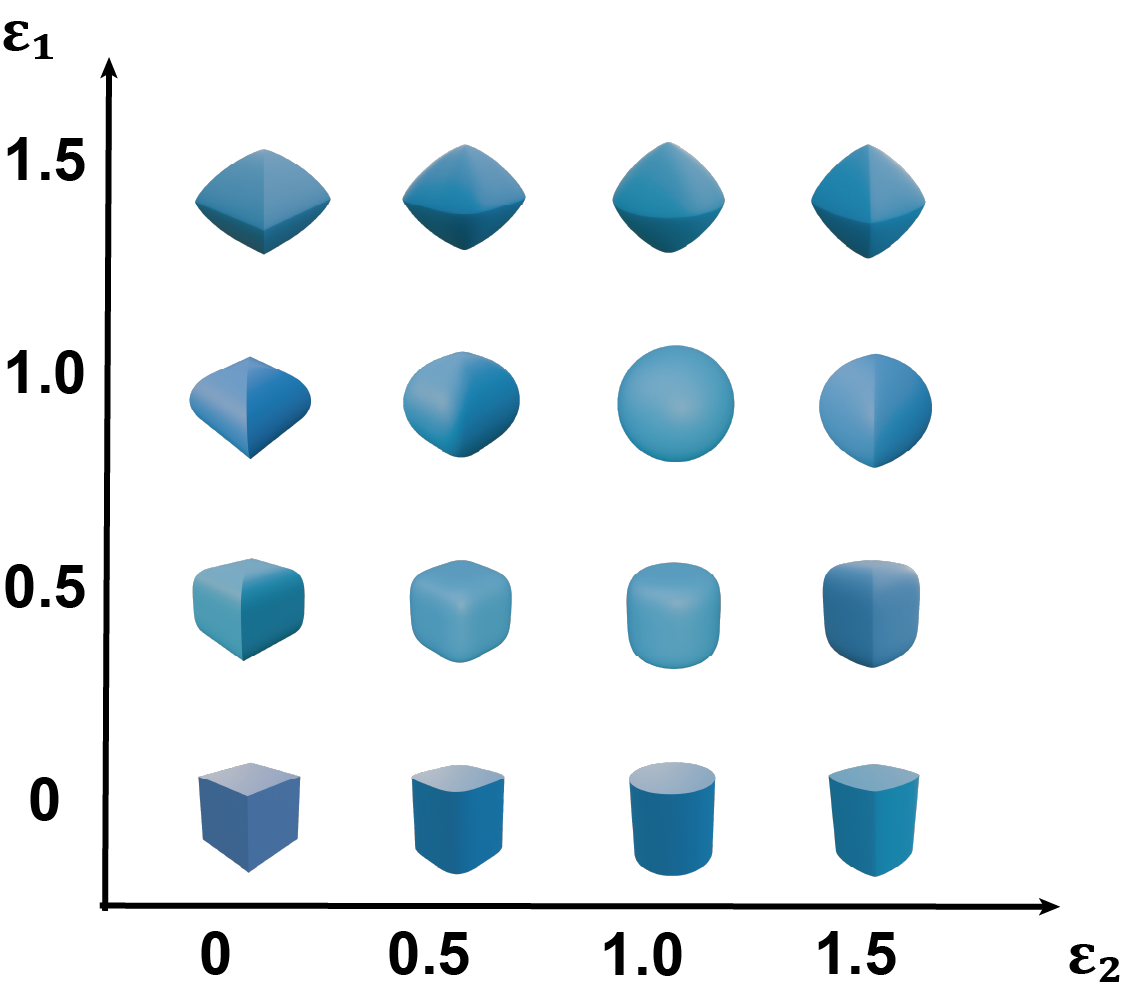}  
    \caption{Shape primitive space for convex superquadrics.}
    \label{fig:super_1}
\end{figure}

Superquadrics are a versatile family of parametric geometric primitives that can represent a wide range of shapes, including cubes, cylinders, spheres, and ellipsoids, within a continuous parameter space \cite{super1}. They are commonly defined using an implicit function, as shown in Equation ~\ref{eq:1}:
\begin{equation}
F(\mathbf{x}) = \left(\left(\frac{x}{a_x}\right)^{\frac{2}{\epsilon_2}} + \left(\frac{y}{a_y}\right)^{\frac{2}{\epsilon_2}}\right)^{\frac{\epsilon_2}{\epsilon_1}} + \left(\frac{z}{a_z}\right)^{\frac{2}{\epsilon_1}} - 1,
\label{eq:1}
\end{equation}
where \(\mathbf{x} = [x, y, z]^T \in \mathbb{R}^3\) is a point in the superquadric’s local coordinate frame. The parameters \(a_x, a_y, a_z > 0\) represent scale factors along the \(x\), \(y\), and \(z\)-axes, respectively. The shape parameters \(\epsilon_1\) and \(\epsilon_2\) (\(\epsilon_1, \epsilon_2 \geq 0\)) control the curvature and convexity of the primitive. When both \(\epsilon_1\) and \(\epsilon_2\) lie in the range \([0, 2]\), the superquadric remains convex, as shown in Figure \ref{fig:super_1}. Superquadrics can also be expressed using an explicit parametric function for surface sampling, as shown in Equation~\ref{eq:superquadric_param}:

\begin{equation}
\mathbf{r}(\eta, \omega) = 
\begin{bmatrix}
  a_x \cos^{\epsilon_1}\eta \cos^{\epsilon_2}\omega \\
  a_y \cos^{\epsilon_1}\eta \sin^{\epsilon_2}\omega \\
  a_z \sin^{\epsilon_1}\eta
\end{bmatrix}
\label{eq:superquadric_param}
\end{equation}

where \(\eta \in [-\pi/2, \pi/2]\) and \(\omega \in [-\pi, \pi]\). This explicit representation facilitates the generation of surface points, enabling efficient use in applications such as shape fitting and surface reconstruction. To fully define a superquadric, a total of 11 parameters are required: three for the size (\(a_x, a_y, a_z\)), two for the shape (\(\epsilon_1, \epsilon_2\)), and six for its pose in 3D space. The pose is represented by \(\mathbf{g} = [\mathbf{R}, \mathbf{t}] \in SE(3)\), where \(\mathbf{R} \in SO(3)\) is the rotation matrix and \(\mathbf{t} \in \mathbb{R}^3\) is the translation vector.

\begin{figure*}[t]
    \centering
    \includegraphics[width=\textwidth]{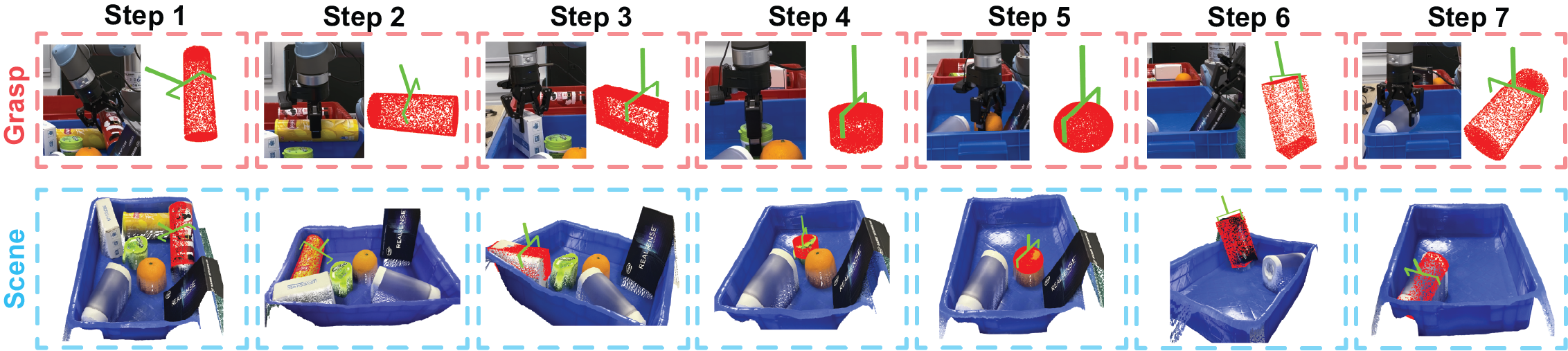}  
    \caption{We illustrate a sequential rollout of the superquadrics-guided robotic grasping process. The red point cloud represents the superquadric fitting for each partial point cloud, while the green vector denotes the grasp sampled from the fitted superquadrics. The top row visualizes the grasping sequence, and the bottom row depicts the evolving scene state after each step.}
    \label{fig:grasping_demo}
\end{figure*}

\subsection{Shape Approximation for Robotic Grasping}
Primitive shape approximation offers a promising approach for robotic grasping as it captures an object’s physical properties and enables generalizable grasp strategies without prior object knowledge. Previous works primarily use Minimum Volume Bounding Boxes (MVBB) \cite{huebner2008, graspit2023} and shape primitive templates \cite{lin2020, Miller2003, access2024}. For instance, \cite{graspit2023} approximates objects using MVBBs and generates grasp candidates based on human demonstrations of grasping similar shapes. While MVBBs may not capture fine object details, they remain effective for grasp sampling and planning. Similarly, \cite{lin2023robot} employs Minimum Oriented Bounding Boxes (OBB) with clustering for part segmentation, simplifying grasp sampling. However, this method frames grasping as a 2D decomposition problem, limiting its generalizability to 3D scenarios. Furthermore, \cite{torii2018} employs a 3D deep neural network to approximate objects as simple primitives (e.g., hexahedrons, cylinders, and spheres) for grasp sampling. While this simplifies grasp planning, the basic primitives lack expressiveness for complex objects. To address this, \cite{lin2020} uses a Mask R-CNN instance segmentation network to detect predefined primitives from depth images, trained on synthetic data and validated on real robots. Extending this, \cite{access2024} expands the primitive template library to include more shapes, improving object representation. However, both shape templates and bounding boxes remain insufficiently expressive for generalizable grasping, highlighting the need for more adaptable shape primitives.

\subsection{Superquadrics for Robotic Grasping}
There has been growing interest in using superquadrics as shape primitives for robotic grasping \cite{{makhal2018}, {wu2023}}. Superquadrics provide a compact parameterization that balances computational efficiency and expressiveness, making them ideal for data-driven approaches \cite{Paschalidou2019}. Early work by \cite{vezzani2017} analyzed superquadric geometry, deriving force-closure and form-closure theories for robotic grasping and formulating grasp pose estimation as a nonlinear optimization problem. This was extended by \cite{vezzani2018}, which integrated prior object knowledge via a shape classifier, reducing computation time and improving stability. Later approaches leveraged mirror symmetry to recover partial point clouds for shape approximation \cite{makhal2018, Huaman2015}, but these methods require object-specific prior knowledge, such as symmetry, limiting their applicability.
A two-stage data-driven method \cite{kim2022} trains two separate networks for part segmentation and primitive fitting. While intuitive, the part segmentation stage relies on relatively complete point clouds, limiting its effectiveness in cluttered scenes \cite{{access2024}}. Furthermore, training two separate networks requires a large amount of data and extensive human annotation for both tasks, raising computational costs.
Subsequent work \cite{SuperQGRASP} uses Neural Radiance Fields (NeRF) to reconstruct complete object shapes and predefine grasps with superquadrics, but NeRF’s reliance on multiple camera viewpoints limits its practicality. Recent work \cite{wu2023} proposes an optimization method to identify hidden superquadrics within object point clouds, emphasizing that local geometry suffices for successful grasps. However, this method relies on nearly complete point clouds, requiring multi-camera setups for point cloud fusion, which is impractical in bin picking.

\section{Methodology}
Our method addresses the challenge of enabling robots to understand cluttered scenes and identify graspable regions from single monocular RGB input in bin-picking tasks. As shown in Figure~\ref{fig:RGBSQGrasp}, our framework processes scenes in multiple phases: \textit{Dataset Generation} (Section~\ref{sec:dataset-generation}) creates a synthetic dataset to simulate occlusions; the \textit{Superquadrics Fitting Network} (Section~\ref{sec:superquadric-fitting}) leverages synthetic data to train a network that extracts superquadric primitives from partial point clouds; during inference,  \textit{SQ-guided Scene Understanding} (Section~\ref{sec:SQ-scene}) estimates the SQ-represented scene using vision foundation models; and \textit{Grasp Pose Sampling} (Section~\ref{sec:grasp-sampling}) generates grasp poses. 


\subsection{Dataset Generation}
The goal of our dataset generation is to create a diverse synthetic dataset using cost-effective simulators. Building on MetaGraspNetV2 \cite{MetaGraspNetV2}, a large-scale dataset based on Isaac Sim \cite{IsaacSim}, we constructed a synthetic primitive dataset by capturing partially observed point clouds for objects with varying occlusion levels in each scene. For each object and its partial point cloud (PCD), the ground truth superquadric (SQ) primitive is sampled from the object model, as shown in Figure~\ref{fig:RGBSQGrasp}. Specifically, for each scene, six camera viewpoints are sampled near the top-down view, including one primary top-down viewpoint and five slightly altered perspectives to introduce variance. For each viewpoint, the segmentation mask separates the point cloud, and the corresponding SQ parameters are associated with each PCD to form a PCD-SQ pair. To enhance generalization, we incorporate data from both the MuJoCo \cite{MuJoCo} and PyBullet \cite{PyBullet} simulators using the same procedure. These simulators introduce variations in rendering effects, such as differences in point cloud density, sparsity, and noise levels, which help reduce the sim-to-real gap and improve the dataset's applicability for real-world tasks \cite{MultiViewFusion}. Our synthetic dataset includes 36K PCD-SQ ground truth pairs for training. To enhance generalization, several data augmentation techniques are applied: Gaussian noise is added to partial point clouds (standard deviation: 0.001–0.005), the point cloud is scaled along each axis within the range [0.5, 2.0] at 0.05 intervals, and the point cloud is randomly translated within the bin's workspace.
\label{sec:dataset-generation}

\subsection{Superquadrics Fitting Network}
In real-world bin-picking scenarios, point clouds often contain missing or occluded data due to the cluttered environments and limited camera views. Superquadric fitting addresses this by approximating visible geometry with generic primitives, making it effective for understanding unknown objects and partial observations. This approach mitigates the limitations of traditional methods, such as point cloud completion, which can be error-prone when large portions of an object are occluded or missing, and typically require prior knowledge of object shapes \cite{scarp2023}, \cite{Fei2022}. Our model is trained to predict superquadric shape parameters from partial point clouds. To prepare the input, the point clouds undergo filtering to remove outlier noise. Subsequently, the point cloud is downsampled to 2000 points using the Farthest Point Sampling (FPS) method \cite{{FPS}}, ensuring efficient and representative sampling of the object’s geometry. The ground truth is generated by uniformly sampling 2000 points from the superquadric surface, generated from ground truth SQ parameters. The network architecture integrates both local and global feature extraction to enhance superquadric fitting accuracy. The local feature extractor utilizes the DGCNN architecture \cite{DynamicGraphCNN}, processing the point cloud through four EdgeConv layers followed by a multi-layer perceptron (MLP). The global feature extractor, adapted from PointNet \cite{pointnet}, predicts scale parameters to capture the object’s overall geometry, which is essential for determining its size and spatial extent \cite{Zhu2024}. Both the local and global feature extractors generate 1024-dimensional permutation-invariant feature vectors. These feature vectors are then passed through an MLP with layers of dimensions (512, 256), followed by leaky ReLU activations. For translation and rotation, the model computes a weighted ensemble of the outputs from both the global and local branches, refining the final predictions by averaging the errors. The model uses four prediction heads: the shape head outputs \( \epsilon_1 \) and \( \epsilon_2 \), the scale head outputs \( \alpha_1 \), \( \alpha_2 \), and \( \alpha_3 \), the rotation head outputs a rotation matrix \( \mathbf{R} \), and the translation head outputs a translation vector \( \mathbf{t} \). These outputs are combined to recover the SQ surface, from which a point cloud of 2000 points is sampled. The training loss compares the predicted point cloud with the ground truth using Chamfer Distance (CD) loss. The CD loss is permutation-invariant, making it suitable for point cloud comparison \cite{{CD_loss}}. This is defined as:
\begin{equation}
\begin{aligned}
d_{\text{CD}}(T_{\text{pred}}, T_{\text{GT}}) &= \frac{1}{|T_{\text{pred}}|} \sum_{t_{\text{pred}_i} \in T_{\text{pred}}} \min_{t_{\text{GT}_j} \in T_{\text{GT}}} \| t_{\text{pred}_i} - t_{\text{GT}_j} \|_2^2 \\
&\quad + \frac{1}{|T_{\text{GT}}|} \sum_{t_{\text{GT}_j} \in T_{\text{GT}}} \min_{t_{\text{pred}_i} \in T_{\text{pred}}} \| t_{\text{GT}_j} - t_{\text{pred}_i} \|_2^2
\end{aligned}
\end{equation}

where \( T_{\text{pred}} \) represents the predicted superquadric (SQ) point cloud, \( t_{\text{pred}_i} \) denotes each point in \( T_{\text{pred}} \), \( T_{\text{GT}} \) represents the ground truth point cloud, and the $|.|$ operator counts the number of elements in the set.
 The loss  $L_{\text{fitting}} = d_{\text{CD}}(T_{\text{pred}}, T_{\text{GT}})$
 ensures alignment of the predicted and ground truth point clouds.

\label{sec:superquadric-fitting}

\begin{figure}[!t]
    \centering
    \includegraphics[width=\columnwidth]{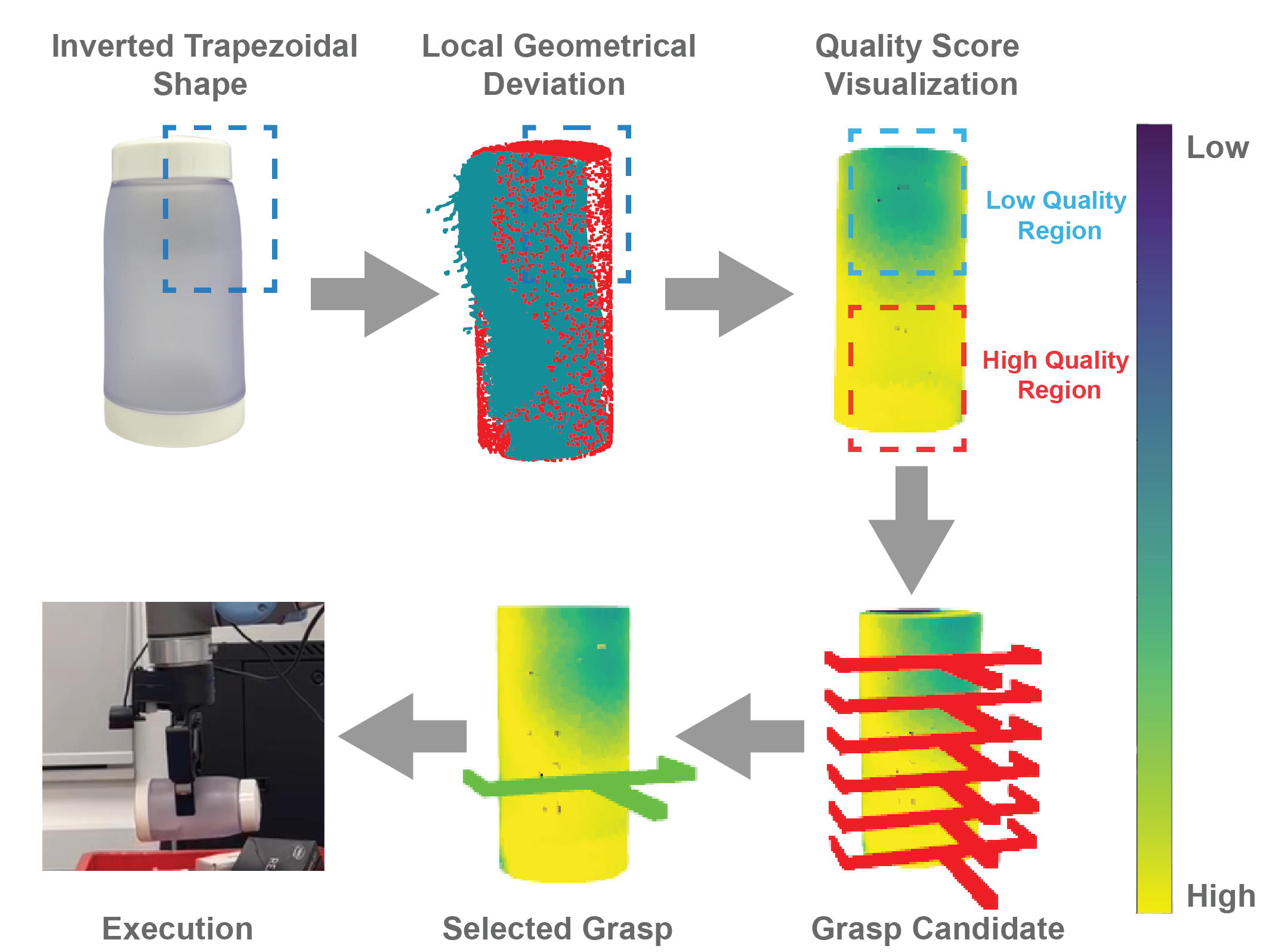}  
\caption{Grasp sampling workflow: Grasp candidates are generated based on superquadric fitting, with selection prioritized from high-quality regions and proximity to the object's center of mass (COM) to ensure stable execution.}
    \label{fig:grasp_workflow}
\end{figure}
\begin{figure}[t]
    \centering
    \includegraphics[width=\columnwidth]{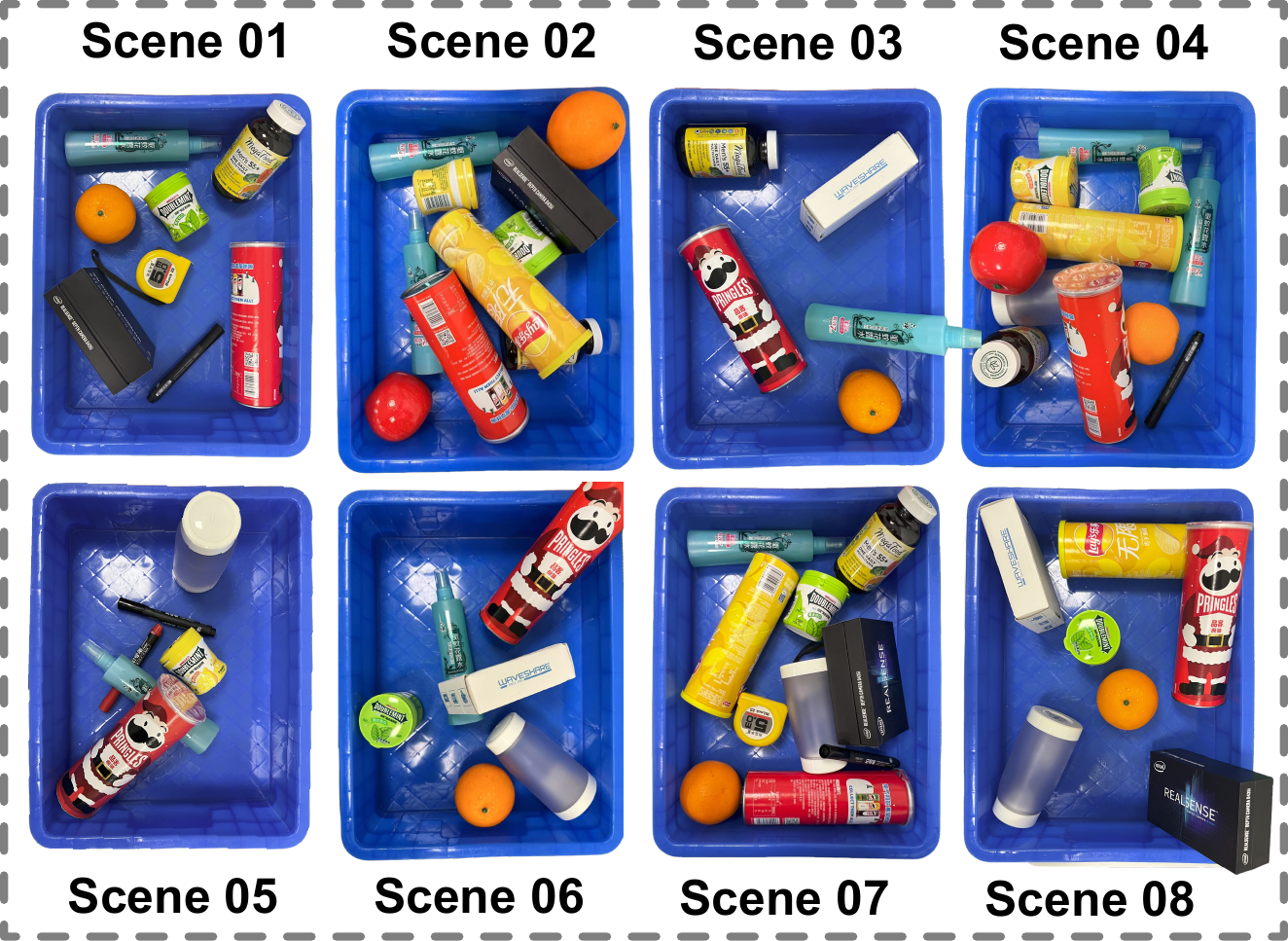}  
    \caption{Example scenes for the real-robot experiments.}
    \label{fig:scene}
\end{figure}

\subsection{SQ-guided Scene Understanding}
For real-robot bin picking deployment, we utilize a Microsoft Azure Kinect RGB camera as a monocular sensor. Prior studies \cite{lin2020, RGBGrasp} have highlighted that depth capture by the sensor often suffers from noise and artifacts, particularly along object edges and reflective surfaces, contributing to inconsistencies between synthetic and real data that degrade model performance. Similarly, RGB data introduce discrepancies due to variations in lighting and color \cite{RGBGrasp}. To mitigate these sim-to-real inconsistencies, we leverage the Depth Pro \cite{depthpro} foundation model for metric depth estimation and Segment Anything 2 \cite{SAM2} for depth-based segmentation. We observe that applying foundation models to RGB images produces depth maps that closely resemble real depth sensor outputs, reducing discrepancies between synthetic and real data and improving generalization. The generated metric depth map provides richer geometric information for shape primitive recognition compared to RGB data \cite{lin2020, MMRNet}, as depth is less affected by texture variations and offers a more transferable representation for unseen objects. By projecting the metric depth onto the segmentation map, we obtain instance-aware partial point cloud (PCD) map. \textbf{Pre-processing Stage.} Each PCD is first preprocessed by downsampling to 2000 points using FPS and applying RANSAC \cite{{RANSAC}} for outlier removal. The preprocessed PCD is then passed through the trained superquadric fitting network to obtain the SQ parameters. \textbf{Post-processing Stage}. This stage further refines the SQ fitting by applying Iterative Closest Point (ICP) alignment. Finally, the SQ point cloud is projected back into the scene for grasp sampling, with each SQ primitive capturing sufficient geometric information for the downstream grasping task.
\label{sec:SQ-scene}

\begin{table*}[t]
\centering
\caption{Performance comparison of grasping and shape primitive recognition for real-robot experiments}
\label{tab:main_results} 
\renewcommand\arraystretch{1.2}
\scalebox{0.86}{
\begin{tabular}{l|cccc|cccc|cccc|cccc}
\hline
\multirow{2}{*}{\textbf{Method}} & \multicolumn{4}{c|}{\textbf{Scene 1}} & \multicolumn{4}{c|}{\textbf{Scene 2}} & \multicolumn{4}{c|}{\textbf{Scene 3}} & \multicolumn{4}{c}{\textbf{Scene 4}} \\ 
& mRE$\downarrow$& mTE$\downarrow$ & mCD$\downarrow$& GSR$\uparrow$ & mRE$\downarrow$& mTE$\downarrow$& mCD$\downarrow$& GSR$\uparrow$& mRE$\downarrow$& mTE$\downarrow$& mCD$\downarrow$& GSR$\uparrow$& mRE$\downarrow$& mTE$\downarrow$& mCD$\downarrow$& GSR$\uparrow$\\ \hline\hline
\textbf{Ours} &  \textbf{5.39}   &   \textbf{1.47}  &  \textbf{13.62}   &  \textbf{95.8\%}   &  \textbf{9.15}   &  \textbf{4.53}   &   \textbf{26.79}  &  \textbf{86.7\%}  &  \textbf{3.27}  &  \textbf{2.46}   &   \textbf{16.48}  &  \textbf{100\%}  &  \textbf{8.35}   &  \textbf{3.66}   &  \textbf{19.51}   &   \textbf{90.1\%} \\ 
PS-CNN \cite{{lin2020}}    &  12.77   &  4.58   &  28.86   &  87.5\% &  21.22   &  8.54   &  163.2   & 76.7\%   &  8.11   & 4.79  &   55.21  &  93.3\%  &  15.89   &  10.21   &  124.97   &  72.7\% \\ 
MMPS  \cite{{access2024}}&  11.24   &   3.14  &   17.85  & 87.5\%   &  12.35   &   8.06  &  94.41   &  80\%  &   7.85  &   5.68  &  68.29   &  86.6\%  &  16.72  &  8.41   &  115.89   &    72.7\%\\ \hline\hline
\multirow{2}{*}{\textbf{Method}} & \multicolumn{4}{c|}{\textbf{Scene 5}} & \multicolumn{4}{c|}{\textbf{Scene 6}} & \multicolumn{4}{c|}{\textbf{Scene 7}} & \multicolumn{4}{c}{\textbf{Scene 8}} \\ 
& mRE$\downarrow$& mTE$\downarrow$& mCD$\downarrow$& GSR$\uparrow$& mRE$\downarrow$& mTE$\downarrow$& mCD$\downarrow$& GSR$\uparrow$& mRE$\downarrow$& mTE$\downarrow$& mCD$\downarrow$& GSR$\uparrow$& mRE$\downarrow$& mTE$\downarrow$& mCD$\downarrow$& GSR$\uparrow$\\ \hline\hline
\textbf{Ours} &  \textbf{8.54}   &  \textbf{2.17}   &  \textbf{29.75}   &  \textbf{88.9\%}  &  \textbf{9.62}   & \textbf{1.96}    &   \textbf{9.15}  &  \textbf{94.4\%}  &   \textbf{3.05}  &  \textbf{2.15}   &  \textbf{8.85}   &  \textbf{92.5\%}  &  \textbf{1.57}   &  \textbf{2.09}   &  \textbf{7.93}   &   \textbf{90.5\%} \\ 
PS-CNN \cite{{lin2020}}  &  15.51   &  10.71   &  148.3   &  72.2\%  &  18.21   &  3.19   &  53.89   &  83.3\%  &  11.58   &  3.45   &   40.76  &   85.2\% &  14.23   &  8.28   &  56.94   &  66.7\%  \\ 
MMPS \cite{{access2024}}&   9.35  &  6.28   &   71.59  &  77.8\%  &  16.46   & 3.65   &   57.26  &  83.3\%  &   7.85  &  8.58   &  24.37   &  85.2\%  &  7.94   &  5.16   &   17.52  &  71.4\%  \\ \hline
\end{tabular}
}
\end{table*}
\begin{table}[t]
\caption{Ablation study results for individual object tasks, reported as the mean across seen and unseen objects}
\label{tab:ablation_results} 
\centering
\renewcommand\arraystretch{1.2}
\resizebox{\columnwidth}{!}{
\begin{tabular}{l|p{0.55cm}<{\centering}p{0.55cm}<{\centering}p{0.55cm}<{\centering}p{0.55cm}<{\centering}|p{0.55cm}<{\centering}p{0.55cm}<{\centering}p{0.55cm}<{\centering}p{0.55cm}<{\centering}}
\hline
\multirow{2}{*}{\textbf{Method}} & \multicolumn{4}{c|}{\textbf{Seen Objects}} & \multicolumn{4}{c}{\textbf{Unseen Objects }} \\ 
& mRE$\downarrow$& mTE$\downarrow$ & mCD$\downarrow$& GSR$\uparrow$ & mRE$\downarrow$& mTE$\downarrow$& mCD$\downarrow$& GSR$\uparrow$\\ 
\hline\hline
\textbf{Ours} & \textbf{0.42}   &   \textbf{0.81}  &  \textbf{4.43}   &  \textbf{100\%}   &  \textbf{1.21}   &  \textbf{2.15}  &   \textbf{8.54}  &  \textbf{94.4\%}  \\ 
Ours w/o Pre    &  2.19   &  1.58   &  5.62   &  91.7\% &  3.12  &  3.51   &  16.73   & 88.9\%   \\ 
Ours w/o PP  &  8.37  &   4.09  &   35.59  & 75\%   &  9.35  &   5.31  &  37.63   &  80.6\%  \\ 
Ours w/o GE  &  1.87   &   2.53  &  9.87   & 91.7\%   &  3.78   &   2.94  &  12.51& 88.9\%  \\ 
Ours w/o LE  &  3.55   &   1.76  &  15.56  & 83.3\%   &  2.81   &   3.87  &  21.19   &  86.1\%  \\ 
\hline
\end{tabular}
}
\end{table}

\subsection{SQ-guided Grasp Pose Sampling}
For each superquadric primitive, a family of grasp pose candidates is sampled from the superquadric shape parameters \(\epsilon_1, \epsilon_2, \alpha_1, \alpha_2, \alpha_3\). The sampling strategy follows the approach proposed in \cite{wu2023}. The shape parameters \(\epsilon_1, \epsilon_2\) define the base grasp type (e.g., rotating around the base or moving vertically along it), while the scale parameters (\(\alpha_1, \alpha_2, \alpha_3\)) define the number of possible grasps along each axis of the superquadric. Key steps include sampling antipodal grasp poses and moving the gripper along specific axes, such as along the principal axes (x, y, and z), to generate a diverse set of candidate poses for each superquadric by leveraging the superquadric’s symmetry and geometric properties.
After generating a set of potential grasp candidates, the grasp pose is further selected based on the following rules:

1. \textbf{Collision Checking}: The generated grasp poses are first collision-checked with the bin and other objects using the Planning Scene module in MoveIt! \cite{{MoveitROS}}.

2. \textbf{SQ Quality Score}: Since fitting an object with a single superquadric may introduce local inaccuracies \cite{wu2023}, a superquadric quality score is computed to assess the fitting reliability. As illustrated in Figure~\ref{fig:grasp_workflow}, certain complex shapes, such as an inverted trapezoidal object, cannot be perfectly represented by a single superquadric, leading to local fitting deviations. These deviations are quantified by the superquadric quality score. After aligning a partial object point cloud \( S \) (source) with the corresponding SQ fitting \( T \) (target), the SQ quality score \( C_i \) for each target point \( t_i \) is calculated based on its distance to the nearest neighbor source point \( s_j \) as:
\begin{equation}
C_i = \frac{\exp(-d_i)}{Z}, \quad d_i = \| t_i - s_j \|
\end{equation}
where \( Z \) is a normalizing constant. Next, the SQ fitting is uniformly sampled into \( N \) point cloud regions, where \( N \) is a parameter based on the size of the SQ fitting. A tunable hyperparameter \( \sigma \) is used to select valid grasp poses. For each region \( R_k \), the average score \( \bar{C}_k \) is computed as:
\begin{equation}
\bar{C}_k = \frac{1}{|R_k|} \sum_{i \in R_k} \frac{\exp(-d_i)}{Z}
\end{equation}
If \( \bar{C}_k > \sigma \), the grasp pose from that region is considered high-quality fitting region with valid SQ fitting.

3. \textbf{Grasp Prioritization}: Within the high-quality region, top-down grasps are prioritized by calculating the dot product \( \mathbf{g} \cdot \mathbf{z} \), where \( \mathbf{z} \) is the unit vector along the positive z-axis, and \(\mathbf{g}\) represents the gripper's approach direction. The grasp selection is based on the \( \text{score} = \mathbf{g} \cdot \mathbf{z} \). Grasps with a score close to 1, indicating alignment with the top-down direction, are favored. If multiple grasps share the same approach direction, the one closest to the center of mass (COM) of the SQ primitive is selected to ensure stability and minimize the risk of imbalance during manipulation \cite{Rodriguez2012}.

4. \textbf{Execution}: The sampled grasp poses are applied to the manipulator, as shown in Figure~\ref{fig:grasping_demo}. A sequential process is executed for each object in the scene until clearance or a predefined maximum number of grasps is achieved.

\label{sec:grasp-sampling}

\section{Experiments and Results}
\subsection{Experimental Setup}
For simulation experiments, PyBullet\cite{{PyBullet}} served as a proof-of-concept platform to validate the feasibility of the RGBSQGrasp framework. We used thirty objects from the YCB \cite{Calli2017} and GraspNet-1B \cite{GraspNet} datasets to generate four randomly designed bin-picking scenes. Objects were freely dropped into the source bin to simulate real-world randomness.

For real-robot experiments, the framework was evaluated on eight randomly designed scenes of varying complexity, as shown in Figure~\ref{fig:scene}. Sixteen distinct objects—four seen and twelve unseen—were selected from a daily supermarket setting to assess its effectiveness. Each scene followed a standard bin-picking setup, where objects were transferred from a source bin to a target bin. To ensure robustness, three repeated experiments were conducted for each scene. Experiments were performed using a UR5e manipulator equipped with a Robotiq 2F85 gripper. Monocular RGB images, captured by a Microsoft Azure Kinect camera, served as input for the framework.

\textbf{Evaluation Metrics.} We evaluate performance using quantitative metrics and real-robot bin-picking outcomes. Following \cite{SuperQGRASP, pcn2018}, superquadric fitting accuracy is measured by mean Chamfer distance (mCD), mean rotational error (mRE), and mean translational error (mTE), which quantify the precision of shape fitting between predicted superquadrics and ground truth. Additionally, the grasp success rate (GSR) is defined as the ratio of successfully picked, transported, and placed objects to the total number of objects in the scene, reflecting system effectiveness. All metrics are computed at the scene level by averaging across all objects, providing a holistic performance measure.

\textbf{Implementation Details.} The framework utilizes pre-trained weights for Depth Pro and SAM2 for metric depth estimation and segmentation. The synthetic dataset is divided into an 80\%/20\% split for training and testing, respectively. The superquadric fitting network is trained on an RTX 4090 GPU for 100K iterations using the Adam optimizer \cite{{kingma2015}} with a learning rate of \(10^{-3}\) and a batch size of 8.

\textbf{Baselines.} To evaluate the effectiveness of \textit{RGBSQGrasp}, we compare it against two baseline methods that employ shape primitives for robotic grasping. The first baseline, PS-CNN \cite{{lin2020}}, predefines shape templates and uses ICP to align segmented point clouds with a template database. To reconcile discrepancies between synthetic training data and real depth captures, PS-CNN incorporates bidirectional image filtering. The second baseline \cite{{access2024}} introduces a multi-modal primitive shape grasping (MMPS) method, which incorporates both RGB and depth modalities to enhance segmentation in packed scenes. 

\begin{figure}
\centering
\includegraphics[scale=0.68]{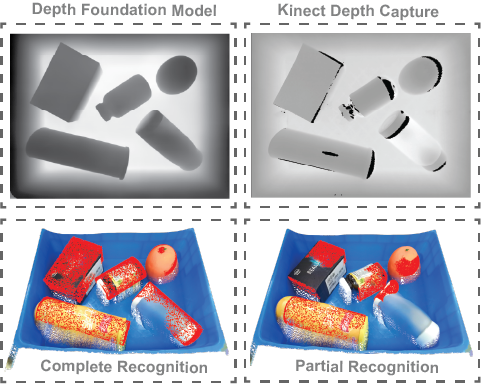} %
\caption{Comparison of depth perception between Depth Foundation Model (left) and Kinect Depth Capture (right). Kinect depth capture introduces noise and artifacts, particularly around object edges and reflective surfaces, leading to incomplete object recognition and affecting scene understanding.}
\label{fig:depth}
\end{figure}

\subsection{Results and Analysis}
The real-robot experiment results, summarized in Table~\ref{tab:main_results}, highlight the superior performance of RGBSQGrasp over PS-CNN and MMPS. RGBSQGrasp achieves an average grasp success rate (GSR) of 92\%, significantly outperforming PS-CNN (79.7\%) and MMPS (80.5\%). A key observation is that in scenes with reflective or translucent objects, such as glass bottles, PS-CNN, which relies heavily on depth data, exhibits a notable performance drop due to corrupted depth capture, as illustrated in Figure~\ref{fig:depth}. While MMPS incorporates RGB modality to partially address this issue, it still struggles with depth inconsistencies. In contrast, RGBSQGrasp leverages a foundation model to enhance sim-to-real consistency, enabling robust primitive recognition and higher grasp success rates even in challenging scenarios. This underscores the effectiveness of RGBSQGrasp in handling reflective and translucent objects, where traditional depth-based methods often fail.

\subsection{Ablation Studies}

We ablate pre-processing ("Ours w/o Pre"), post-processing ("Ours w/o PP"), the global encoder ("Ours w/o GE"), and the local encoder ("Ours w/o LE") to quantify their contributions to the framework's performance. Our ablation study utilizes the dataset from real-robot bin picking experiments, consisting of sixteen objects (four seen, twelve unseen). We evaluate superquadric fitting accuracy and robotic grasping performance through object-specific tasks, isolating individual objects to avoid metric distortion caused by clutter.

The results in Table~\ref{tab:ablation_results} demonstrate the effectiveness of each component in our framework. The full model ("Ours") achieves the best performance across all metrics, with the lowest mRE, mTE, and mCD, and the highest GSR for both seen and unseen objects. A key observation is that combining global and local features yields superior results: global features excel at estimating overall geometric scale, while local features capture fine-grained details. Removing either the global encoder ("Ours w/o GE") or local encoder ("Ours w/o LE") degrades performance, particularly in mCD and GSR. Additionally, post-processing refinement ("Ours w/o PP") is critical, as it aligns the estimated superquadric shape with the partial point cloud, enhancing recognition accuracy and grasping sampling.

\section{CONCLUSIONS}

In this work, we propose a superquadric-guided framework for bin picking using a single monocular RGB image, enabling robust scene understanding and grasp synthesis without CAD models or object priors. Real-robot experiments demonstrate its effectiveness, achieving a 92\% grasp success rate over 189 picks in varying occlusion scenes, outperforming baseline methods reliant on real depth sensor data. Future work will explore extending the framework to other end-effectors, such as suction cups and robotic hands, as the graspability analysis enabled by superquadrics is generalizable.
\addtolength{\textheight}{-1cm}   





\end{document}